\documentclass[10pt,twocolumn,letterpaper]{article}
\pdfoutput=1
\usepackage{cvpr}
\usepackage{times}
\usepackage{epsfig}
\usepackage{graphicx}
\usepackage{amsmath}
\usepackage{amssymb}
\usepackage{algorithm}
\usepackage{algorithmic}
\usepackage{breqn}
\usepackage{tabularx,booktabs}
\usepackage{bm}
\usepackage{multirow}
\usepackage{soul}

\usepackage[breaklinks=true,bookmarks=false]{hyperref}

\cvprfinalcopy 

\def\cvprPaperID{****} 

\setstcolor{red}
\begin{document}
	\renewcommand{\st}{\vphantom}
	\title{Sparse Label Smoothing Regularization for Person Re-Identification}
	
	\author{Jean-Paul Ainam\thanks{Jean-Paul Ainam is a Ph.D student at the University of Electronic Science and Technology of China. He is also a Lecturer at Adventist Cosendai University, Cameroon.} \\
		{\tt\small jpainam@uacosendai-edu.net}
		\and Ke Qin\thanks{Corresponding author.} \\
		{\tt\small qinke@uestc.edu.cn}
		 \and Guisong Liu\thanks{This author is an equal corresponding author.} \\
		{\tt \small lgs@uestc.edu.cn}
		\and Guangchun Luo \\
		{\tt\small gcluo@uestc.edu.cn}\\
		School of Computer Science and Engineering \\
		University of Electronic Science and Technology of China \\
		Chengdu, Sichuan, P.R. China, 611731 \\
	}
	
	\maketitle

	\begin{abstract}
Person re-identification (re-id) is a cross-camera retrieval task which establishes a correspondence between images of a person from multiple cameras.
Deep Learning methods have been successfully applied to this problem and have achieved impressive results. However, these methods require a large amount of labeled training data. Currently labeled datasets in person re-id are limited in  their scale and manual acquisition of such large-scale datasets from surveillance cameras is a tedious and labor-intensive task.
In this paper, we propose a framework that performs intelligent data augmentation and assigns partial smoothing label\st{smoothing} to generated data. Our approach first exploits the clustering property of existing person re-id datasets to create groups of similar objects that model cross-view variations. Each group is then used to generate realistic images through adversarial training. Our aim is to emphasize feature similarity between generated samples and the original samples. 
Finally, we\st{finally} assign\st{ed} a non-uniform label distribution to the generated samples and \st{defines} define a regularized loss function \st{and a smoothing regularization term to train} for training. The proposed approach tackles two problems (1) how to efficiently use the generated data and (2) how to address the over-smoothness problem found in current regularization methods. 
Extensive experiments on four large-scale datasets show that our regularization method significantly improves the Re-ID accuracy compared to existing methods.
	\end{abstract}
	
\section{Introduction}
	Person re-identification is the problem of identifying persons across images using different cameras or across time using a single camera. \st{Given a person query image, person re-id searches the person images with the same identity from a large gallery that contains person identities observed by another cameras.} Automatic person re-id has become essential in surveillance systems due to the rapid expansion of large-scale and distributed multi-camera systems. However, many issues such as view point variations, dramatic variations in visual appearance, unstable light conditions, human pose variations, clothing similarity, background clutter and occlusions still prevent the task of achieving high accuracy. Despite the increasing attention given by researchers to solve the person re-id problem, it has remained a challenging task in practical environments.

Current approaches to solving person re-id are based on Convolutional Neural Network (CNN) and generally follow a verification or identification framework.
A verification framework \cite{Varior2016Gated, Li2014Pairing, zheng2016discriminatively} usually takes a pair of images as input and outputs a similarity score while an identification framework \cite{Li2018Harmonious, Rahimpour2017Attention, Wu2018CoAttention, zheng2017unlabeled} learns a robust and discriminative feature representation from a single input image and \st{applies a multi-class classification task to} predicts the person identity.

In general, CNN-based approaches to person \st{re-identification} re-id task received remarkable improvements and presented potentials for practical usage in modern surveillance system. 
However, CNN based methods require a large volume of labeled data for training to generalize. Furthermore, existing labeled datasets in person re-identification are limited in their scale by the number of the training images and by the number of images available for each identity. For example, \st{on}Market-1501 dataset \cite{Zheng2015} contains $12,936$ training images and $751$ identities, with $17$ images on average per identities (i.e. $12,936/751$).
Moreover, the need of large datasets becomes obvious as the task of labeling is manual, particularly tedious and labor-intensive. In addition, it involves manual selection of identities and association of images from different cameras with various view points, illumination, occlusions and body pose changes. This lack of large datasets is a big challenge in applying deep learning techniques to person re-id. Therefore, it is very important to find intelligent way to increase the training set.

Recently, Generative Adversarial Networks (GAN) \cite{Goodfellow2014GAN} models have
been particularly popular due to their ability to generate realistic-looking images via adversarial training. Thus, they can be used to solve the problem of lack of large datasets by generating synthesized unlabeled images which can be used in conjunction with the training set. However, transferring unlabeled images from the generated set to the training set is a challenging task and remains unresolved. Early studies to solve this problem adopted simplistic approaches. For instance, "All in one" \cite{Salimans2016ImprovedTF} method assigns a single new label i.e., $K + 1$, to every generated sample. And, "Pseudo Label" \cite{Lee2013PseudoLabel} assigns the maximum class probability predictions of a pre-trained CNN model to the generated sample.
Similarly, \cite{Huang2018MPRL, zheng2017unlabeled, zhong2018camera} proposed to use Label Smooth Regularization (LSR) to assign labels to fake samples. LSR was proposed in the 1980s and recently revisited in \cite{Szegedy2016} as a mechanism to reduce over-fitting by estimating a marginalized effect over non-ground truth labels $y$ during training by assigning small value to $y$ instead of $0$. Specifically, \cite{zheng2017unlabeled} extends LSR to outliers (LSRO) by assigning uniform label distribution (i.e. $\frac{1}{K}$) to generated images. This choice was made to avoid classifying generated samples into one of the existing categories. However, we argue that generated images have considerable visual differences and assigning same labels to all would lead to ambiguous predictions. This claim is also supported by \cite{Huang2018MPRL}.
Along this line, \cite{Huang2018MPRL} proposed to assign labels based on the normalized class predictions over all pre-defined classes. We find that \cite{Huang2018MPRL}'s method is similar to "Pseudo label" \cite{Lee2013PseudoLabel} and besides, empirical experiments conducted by \cite{zheng2017unlabeled} showed that LSRO is superior to "All in one" and "Pseudo-label". 

One major drawback of all existing LSR approaches such as LSRO is that, they can easily lead to over-smoothness especially when the number of classes is excessively large. For instance, in a practical environment with thousand of identities, uniform label smoothing approach will assign value close to $0$ and will fail to model the underlying relationships between the labeled and unlabeled data samples. 
In this work, we attempt to overcome this shortcoming by dynamically associating unlabeled samples with a subset of the class label distribution during the training process. Inspired by clustering that leverages the underlying patterns within data, we propose a novel label assigning approach called Sparse Label Smoothing Regularization (SLSR) which delivers significant performance boost in person re-identification, specifically for large-scale dataset. 

In this paper, we make the following contributions:
\begin{enumerate}
	\item We propose a GAN-based model tailored for person re-identification task with Sparse Label Smoothing Regularization (SLSR).
	\item We use k-means to do clustering on the training set, generate GAN-based samples for each cluster and use partial smoothing label regularization over the generated images.
	\item Using extensive experiments, we show that feature representation learning with SLSR improves the person re-identification accuracy.
\end{enumerate}
The rest of this paper is organized as follows. Section \ref{retaledworks} surveys the related works in person re-identification. Section \ref{modeling} presents the proposed regularization method. Section \ref{frameworkoverview} presents the framework architecture; section \ref{experiments} shows the implementation details and the experimental results and section \ref{conclusion} concludes the paper.

\section{Related works} \label{retaledworks}
In this section, we describe the works relevant to our pipeline. These works include person re-identification and Generative Adversarial Network.
\subsection{Person Re-Identification}
Related works in person re-id can be roughly divided into two groups: distance metric learning and deep machine learning based approaches. The first group, also known as discriminative distance metric focuses on learning local and global feature similarities by leveraging inter-personal and intra-personal distances \cite{ Chen2016, Kostinger2012, Liao2015, Xiong2014, Zhang2011, Zheng2015}. The second group is CNN-based with a goal to jointly learn the best feature representation and a distance metric.
 Some feature based learning approaches  \cite{Cheng2016, Li2017DeepContext, Sun2017RPP} first decompose the images into three parts. Each part is then passed into a number of sub-networks for feature extraction. The three parts are finally fused at the fully connected layers and jointly contribute to the training process using a triplet loss function.
 Other methods \cite{Li2014Pairing, Varior2016Gated, zheng2016discriminatively} used a Siamese convolutional neural network architecture for simultaneously learning a discriminative feature and a similarity metric. Given a pair of input images, they predict if it belongs to the same subject or not through a similarity score.
  To improve the similarity score, \cite{Paisitkriangkrai2015, Zhong2017reranking} proposed to optimize the evaluation metrics commonly used in person re-id.
  
   Recently, \cite{Zhang2018Crossing, zheng2017unlabeled, zhong2018camera} proposed to address the problem of lack of large datasets in person re-id by training a GAN \cite{Goodfellow2014GAN} model to generate samples and a CNN model for identification task. It was particularly observed that, generated images with smooth labels can improve person re-id accuracy when they are combined with the training samples.

Following the success of attention mechanisms in Natural Language Processing, \cite{Liu2017End2End, Li2018Harmonious, Rahimpour2017Attention, Wu2018CoAttention} explored its application to the person re-id problem by proposing various forms of attentions. In details, \cite{Liu2017End2End} proposed an end-to-end Comparative Attention Network (CAN) to progressively compare the appearance of a pair of images and determine whether the pair belongs to the same person. During training, a triplet of raw images is fed into CAN for discriminative feature learning and local comparative visual attention generation. \cite{Li2018Harmonious} proposed a CNN architecture for jointly learning soft and hard attention. The two attention mechanisms with feature representation learning are simultaneously optimized.
 In addition, \cite{Rahimpour2017Attention} proposed gradient-based attention mechanism to solve the problem of pose and illumination found in person re-id problem in a triplet architecture and \cite{Wu2018CoAttention}  recommended Co-attention based comparator to learn a co-dependent feature of an image pair by attending to distinct regions relative to each pair.
  \cite{zheng2016discriminatively} proposed a Siamese network with verification loss and identification loss and predicted the identities of a pair of input images.

  Many semi-supervised and unsupervised methods based on GAN have been developed \cite{Barros2018, yu2017cross, Zhang2018Crossing, zhong2018camera}  to address the problem of lack of large labeled dataset in person re-id. \cite{Barros2018} introduced, for the first time in the re-id field, the strategy of using synthetic data as a proxy for the real data and claim to recognize people independently of their clothing.
 \cite{zheng2017unlabeled} showed that a regularized method (LSRO) over GAN-generated data can improve the person re-id accuracy by assigning uniform label distribution to generated samples. \cite{zhong2018camera} proposed a camera style (CamStyle) adaptation method to regularize CNN training through the adoption of LSR and used CycleGAN \cite{Zhu2017CycleGAN} for image generation.
Similarly, \cite{Lee2013PseudoLabel} trained a supervised network with labeled and unlabeled data by assigning \textit{pseudo-label} to unlabeled data and \cite{Wang2017, yu2017cross} proposed unsupervised asymmetric metric learning to unsupervised person re-id. In addition, \cite{Papandreou2015} proposed Expectation-Maximization (EM) combining weak and strong labels under supervised and semi-supervised settings for image segmentation.
\cite{Li2018RegionMetric} proposed a semi-supervised region metric learning method to improve the person re-id task performance under imbalanced unlabeled data using label propagation with cross person score distribution alignment and discriminative region-to-region metric. Recently, \cite{Minxian2018Tracklet} proposed a domain adaptation method to address the problem of lack of exhaustive identity label. Their proposed model jointly learns per-camera tracklet association and cross-camera tracklet correlation by maximising the discovery of tracklet across camera views and by exploiting the underlying re-id discriminative information in an end-to-end optimization.

Building from \cite{Radford2015, Szegedy2016, zheng2017unlabeled, Zhu2017CycleGAN}, we propose a label assignment strategy that assigns partial label distribution to generated samples. We intend to use the training data in conjunction with GAN generated images to train the network using a regularized loss function.

 We show in section \ref{discussion} how our model differs from \cite{zheng2017unlabeled} and \cite{zhong2018camera}.
\subsection{Generative Adversarial Network}
\begin{figure*}
	\centering
	\includegraphics[scale=0.5]{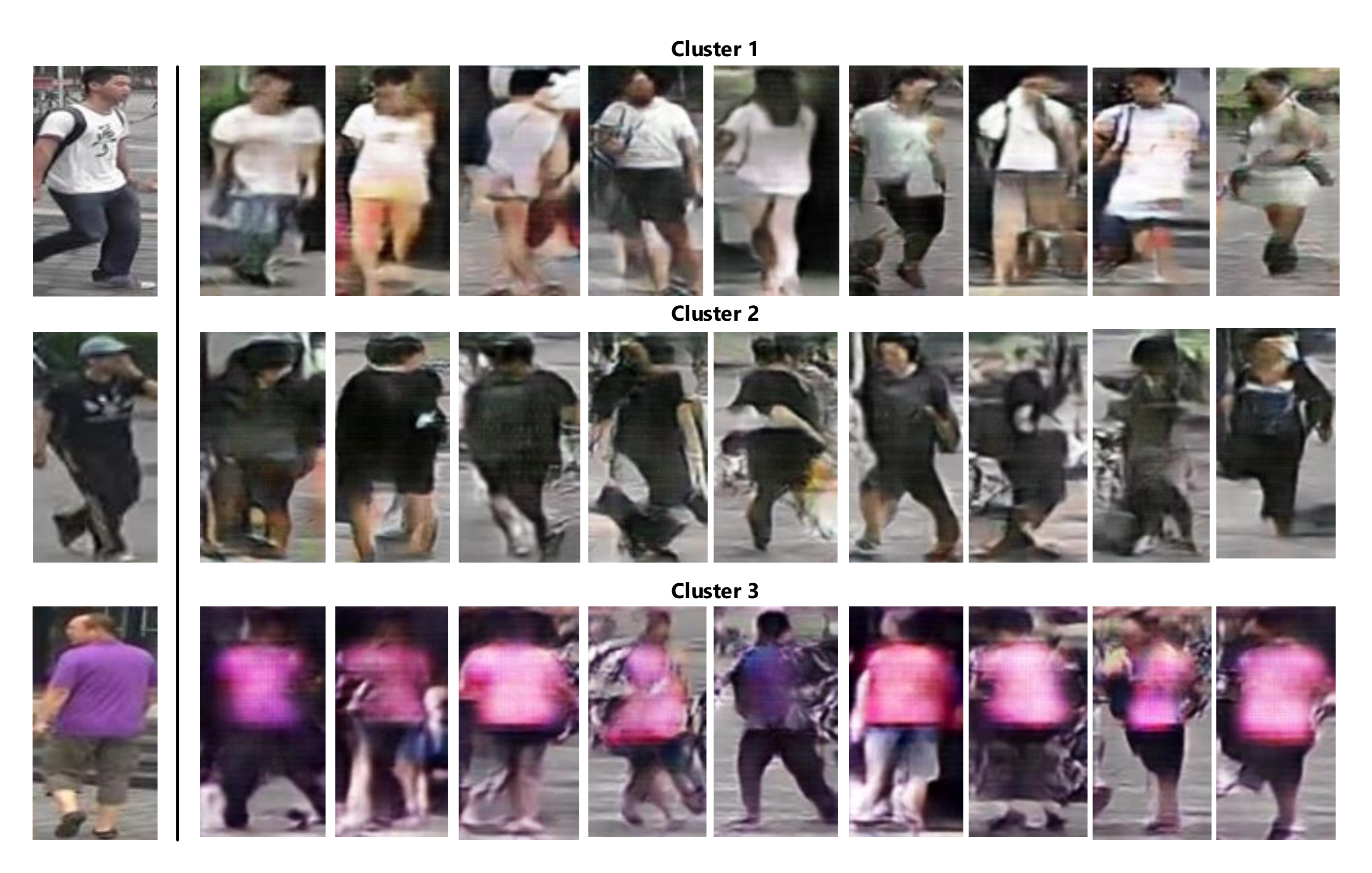}
	\caption{Sample images generated from three clusters using DCGAN. The first column shows the original images from the cluster set and the remaining columns show samples generated from the corresponding cluster. We show that identities with similar features also generate fake samples with similar features and that color is a major learned feature.}
	\label{fig:sampleimages}
\end{figure*}

Generative Adversarial Network (GAN) is first introduced by \cite{Goodfellow2014GAN} and \st{is}described as a framework for estimating generative models via an adversarial process. GAN consists of two different components: a generator (G) that generates an image and a Discriminator (D) that discriminates real images from generated images. The two networks compete following the minimax two-player game.  This kind of learning is called Adversarial Learning. \cite{Radford2015} proposed Deep Convolutional GAN (DCGAN) and certain techniques to improve the stability of GANs. The trained DCGAN showed competitive performance over unsupervised algorithms for image classification tasks.
 Multiple variants of GANs were published in the literature and were applied to various interesting tasks such as realistic image generation \cite{Radford2015}, text-to-image generation \cite{Reed2016}; video generation \cite{Vondrick2016}; image-to-image generation \cite{Isola2016}, image inpainting \cite{Pathak2016}, super-resolution \cite{Ledig2016} and many more. In this work, we use DCGAN \cite{Radford2015} model to generate unlabeled images for each cluster set. We chose DCGAN model after carefully contrasting various image generators. DCGAN architecture is very simple but yet generates more realistic images as illustrated in Figure \ref{fig:sampleimages}.


\section{Our Approach} \label{modeling}
In this section, we present our proposed framework.
\subsection{Clustering the Training set}
We intend to partition the training samples into $K$ groups of equal variance and find a shared feature space among similar objects. Our goal is to produce $K$ different clusters with relatively similar features. To do this, we defined an objective function like that of \textit{k-means} clustering \cite{Aljalbout2018, Hartigan1975}.
\begin{equation}
\label{equation:kmeansobjective}
\mathcal{L}_{clustering} = \sum_{i=1}^{N}\sum_{k=1}^{K} \mid\mid z_i - \mu_k \mid\mid^2
\end{equation}
where $N$ is the number of cases, $\mu_k$ the cluster center and $\mid\mid.\mid\mid $ the Euclidean distance between an embedded data $z_i$ and the cluster center $\mu_k$. In our experiments, we replaced $z_i$ by the output feature map produced by a pre-trained model.
 Equation \ref{equation:kmeansobjective} learns the centroid such that, given a threshold $\gamma$, distances between similar feature vector are smaller than $\gamma$, while those between dissimilar feature vector are greater than $\gamma$. This ensures that distance between generated samples and a subset of the training images is small. We argue that using a generative model on similar objects effectively contributes in maintaining the complex relationships between unlabeled and labeled data, minimizes the affinity distance between the two sample sets and approximates the actual training data.
  In addition, experimental results have shown that using the intermediary feature representation of a pre-trained CNN model instead of the raw image results in better clustering quality.

To generate realistic images from each cluster, we defined a loss function similar to \cite{Goodfellow-et-al-2016} and minimized Equation \ref{equation:ganloss} with respect to the parameters of $G(z)$ and maximized Equation \ref{equation:ganloss} with respect to the parameters of $D(x)$.
\begin{equation}
\mathcal{L}_{GAN}=\log D(x) + \log \Big(1-D(G(z))\Big)
\label{equation:ganloss}
\end{equation}
\subsection{Sparse Label Distribution Scheme}
Let $p(\tilde{y}_i=y_i\vert \bm{I}_i)$ be a vector class probabilities produced by the neural network for an input image $\bm{I}_i$ and  $\bm{w}_i$ the combination of weight and bias terms to be learned\st{for label $y_i$}. The network computes the probabilities of each input image using:
\begin{equation}
p(\tilde{y}_i=y_i\vert \bm{I}_i)=\frac{\text{exp}({\bm{w}^T_{y_i} \cdot \bm{x}_i})}{\sum_{k=1}^{\mathit{N}}\text{exp}({\bm{w}^T_k} \cdot \bm{x}_i)}
\end{equation}
where $\bm{x}_i$ is the input vector from previous layers. Given  $\mathit{N}$ training samples, we define the cost function for real images as the negative log-likelihood:
\begin{equation}
\mathcal{L}_{xent} = -\sum_{i=1}^{N}\log p(\tilde{y}_i = y_i | \bm{I}_i)
\label{equation:entropy1}
\end{equation}
In general, neural network represents a function $f(x;\theta)$ which provides the parameters $\bm{w}$ for a distribution over $y$. So minimizing $\mathcal{L}_{xent}$ is equivalent to maximizing the probability of the ground-truth label $p(\tilde{y}_i = y_i \vert \bm{I}_i)$. For a given person with identity $y$, Equation \ref{equation:entropy1} can be written as
\begin{equation}
\mathcal{L}_{xent}(\theta) = -\log p(y | \bm{x}; \theta)
\label{equation:entropy2}
\end{equation}
where $\bm{\theta}$ represents  the set of parameters of the whole network to be learned.

\textbf{Regularization via Sparse Label Smoothing (SLSR)}
\cite{Szegedy2016} proposed a mechanism to regularize a classifier by estimating a marginalized effect over non-ground truth labels $q(k \vert x)$ during training by assigning small value to $y$ instead of $0$. $q(k \vert x) = \delta_{k,y}$ where $\delta_{k,y}$ is Dirac delta:
\begin{equation} \label{eq:dirac}
\delta_{k,y} = \left\{
\begin{array}{ll}
1 \quad  k = y\\
0 \quad  k \neq y
\end{array}
\right.
\end{equation}
For training image with ground-truth label $y$, \cite{Szegedy2016} replaced the label distribution $q(k\vert x) = \delta_{k,y}$ with
\begin{equation} \label{eq:lsr}
q\prime (k,y) = \left\{
\begin{array}{ll}
(1 - \epsilon) \delta_{k,y} \qquad  k = y\\
\frac{\epsilon}{k} \qquad \qquad \qquad k \neq y
\end{array}
\right.
\end{equation}
where $\epsilon \in [0, 1]$ is the smoothing parameter. When $\epsilon = 0$, Equation \ref{eq:lsr} can be reduced to Equation \ref{eq:dirac}. Then, the cross-entropy loss in Equation \ref{equation:entropy2} is re-defined as
\begin{equation}
\mathcal{L}_{LSR} = - (1-\epsilon)\log p(y \vert x;\theta) - \frac{\epsilon}{K}\sum_{i=1}^{K}\log p(y_i \vert x; \theta)
\end{equation}

Departing from \cite{Szegedy2016}, we introduce our loss function for the feature representation learning as a combination of cross entropy and a modified version of LSR. Given an identity I
\begin{equation}
z_{i,c} = \left\{
\begin{array}{ll}
1 \quad  \bm{I}_i \in \mathcal{C}\\
0 \quad \bm{I}_i \notin \mathcal{C}
\end{array}
\right.
\end{equation}
Here, $z_{i,c}$ are the unnormalized probabilities of an image generated using cluster $\mathcal{C}$ with $p_c$ number of classes. $z_{i}$ represents a one-hot encoding vector where every entry $k$ is equal to $1$ if the class label $k$ belongs to $\mathcal{C}$ and $0$ if not. We consider the ground-truth distribution over the generated image and normalize  $z_i$ so that $\sum_{i=1}^{N}z_{i,c} = 1$.  To explicitly take into account our label regularization, we changed the network to produce
\begin{equation}
z_i = \frac{1}{p_c}z_{i,c} \qquad \text{for} \quad c \in \{1,2, \ldots, K\}
\end{equation}
Figure \ref{fig:regularization} illustrates our proposed label distribution scheme. We finally optimize $\sum_{i} \mathcal{L}(\tilde{z}_i, \frac{1}{p_c}z_{i,c})$.  
Our loss for generated images is written as:
\begin{equation}
\mathcal{L}_{SLS} = - \sum_{i=1}^{p_c}\log p(\tilde{z}_i = z_i \vert \bm{I}_i)
\label{equation:sls1}
\end{equation}
or simply written as
\begin{equation}
\mathcal{L}_{SLS}(\theta) = -\log(p(z \vert \bm{x}; \theta)
\label{equation:sls2}
\end{equation}
Combining Equation \ref{equation:entropy2} and Equation \ref{equation:sls2}, the proposed regularized loss function $\mathcal{L}_{SLSR}$ is defined as:
\begin{equation}
\mathcal{L}_{SLSR}(\theta) = -(1-\lambda)\log \Big(p(y \vert \bm{x}; \theta) \Big)  - \frac{\lambda}{K}\log\Big(p(z \vert \bm{x}; \theta)\Big)
\end{equation}
For training images, we set $\lambda = 0$ and for the generated images, $\lambda = 1$
\subsection{Discussion} \label{discussion}
Recently, \cite{zheng2017unlabeled} proposed Label Smoothing Regularization for Outliers (LSRO) and  \cite{zhong2018camera} proposed CamStyle as a data augmentation technique. LSRO expands the training set with unlabeled samples generated by DCGAN \cite{Radford2015} and assigns uniform LSR \cite{Szegedy2016} to a generated sample i.e. $\mathcal{L}_{LSR}(\epsilon = 1)$ while CamStyle uses CycleGAN \cite{Zhu2017CycleGAN} to generate new training samples according to camera styles and assigns $\mathcal{L}_{LSR}(\epsilon = 0.1)$ to style-transferred images. Although LSRO and CamStyle are similar to our work, we argue that our method is different on two aspects:

{\bf 1)} LSRO \cite{zheng2017unlabeled} and CamStyle \cite{zhong2018camera} assign equal smoothing label distribution to all generated images; this can lead\st{s} to over-smoothness especially when the number of classes is excessively large.
However, our method assigns an adaptive smoothing label distribution to a generated sample based on the label distribution of its cluster $c$ i.e $\mathcal{L}_{LSR}(\epsilon = \frac{1}{p_c})$ where $p_c$ is the number of class identity in cluster $c$. In SLSR, $\epsilon = \frac{1}{p_c}$ is not unique and depends on $p_c$. 
This is opposed to $\epsilon = 1$ and $\epsilon = 0.1$ used in LSRO and CamStyle, respectively. Moreover, in LSRO and CamStyle, dissimilar and similar images may be assigned relatively equal similarity value, while our method deals with such unfairness by considering a generated image in the locality of real samples and proposes a strategy to determine the appropriate candidates by using \textit{k-means} clustering algorithm. A non-uniform label distribution is assigned to generated images according to their cluster of origin. This enables our model to be highly efficient in dealing with large amount of data while being robust to noise as well. Our method SLSR learns the most discriminative features and can easily avoid the over-smoothness problem.

\begin{figure*}
	\centering
	\includegraphics[scale=0.6]{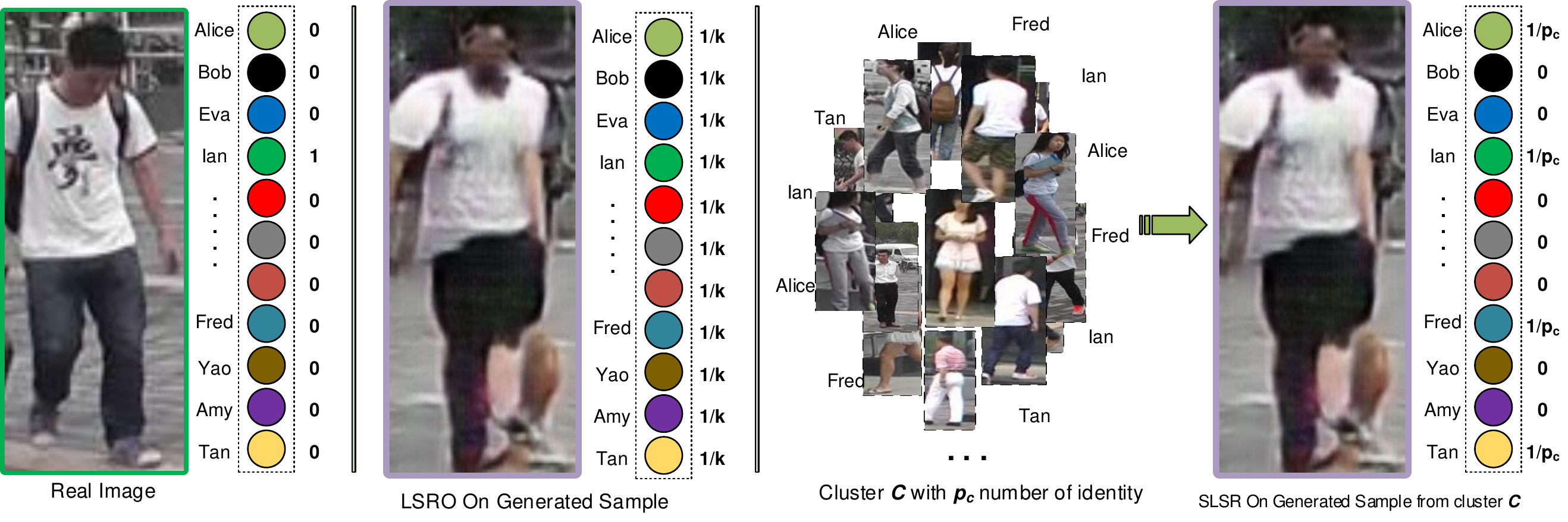}
	\caption{Real image (left) uses one-hot vector to encode the label information. LSRO (middle) uses a uniform label distribution $\frac{1}{k}$ on generated samples, while SLSR (right) uses partial label distribution drawn from the label distribution of the cluster of origin for label information.}
	\label{fig:regularization}
\end{figure*}

{\bf 2)} In our model, similarities are maintained and propagated through the framework by the concatenation of similar images into one homogeneous feature space. Leveraging feature space for each cluster can substantially improve the performance of person re-identification compared with using single-label distribution over all classes. Figure \ref{fig:regularization} illustrates the label distribution of SLSR and LSRO and clearly describes the uniform distribution of LSRO versus the non-uniform distribution of SLSR. Comparative studies in Tables \ref{tab:cuhk03results} \ref{tab:dukeresults} \ref{tab:marketresults} \ref{tab:viperresults} ascertain\st{s} the effectiveness of our method and extensive experiments demonstrate its superiority compared to LSRO \cite{zheng2017unlabeled} and CamStyle \cite{zhong2018camera}.
In addition, our framework introduces an extra noise layer to match the noisy GAN label distribution. The parameters of this linear layer can be estimated as part of the training process and involve simple modification of current deep network architectures.

LSRO, CamStyle and our method SLSR share some common practices such as (1) enhancing the training set by the generation of fake images using GAN \cite{Goodfellow2014GAN} models; (2) the adoption of Label Smooth Regularization (LSR) proposed by \cite{Szegedy2016} to alleviate the impact of noise introduced by the generated images; (3) performing an end-to-end training for person re-id using labeled and unlabeled data in a CNN-based approach.

\begin{table*}
	\caption{Properties comparison between LSRO, All-in-One, Pseudo Label and our method (SLSR)}
	\centering
	\begin{tabular}{lcccc}
		\hline
		Methods & Label distribution & Label contribution & Label source & Label assignment \\
		\hline\hline
		All-in-One \cite{Salimans2016ImprovedTF} & One Hot Encoding & Same & Manual & Static \\
		Pseudo Label \cite{Lee2013PseudoLabel} & One Hot Encoding & Different & Probability & Dynamic \\
		LSRO \cite{zheng2017unlabeled} & Smooth Encoding & Same & Manual & Static \\
		SLSR & \textbf{Smooth Vector} & \textbf{Different} & \textbf{Similarity} & \textbf{Dynamic}\\
		\hline
	\end{tabular}
	\label{tab:slsrproperties}
\end{table*}
We also compared SLSR properties with LSRO, "Pseudo Label" and "All-in-one" methods. 
\begin{algorithm}[t] \label{slsralgorithm}
	\caption{Algorithm for SLSR Training}
	\begin{algorithmic}[1]
		\renewcommand{\algorithmicrequire}{\textbf{Input:}}
		\renewcommand{\algorithmicensure}{\textbf{Output:}}
		\REQUIRE $\mathcal{K}$: Number of clusters, $\mathcal{X}$: Training samples
		\\ \textit{Initialisation}: Randomly initialize the cluster centroids $\mu_1,\mu_2, \ldots, \mu_k \in \mathbb{R}^n$ \st{randomly}
		\\ \STATE Draw \textit{m} samples $ \{(x^{(1)},y^{(1)}), \ldots, (x^{(m)}, y^{(m)}\}  $ from the training data $\mathcal{X}$ and train a CNN for {\bf I} iteration using Equation \ref{equation:entropy2}
		\\
		\FOR {{\bf each} sample \textit{m}}
		\STATE Extract $x^{(n)}_{(m)}$ feature map from the last conv layer
		\ENDFOR
		\STATE Let $\mathcal{F} \in \mathbb{R}^{N\times M}$ be the feature maps for all samples
		\REPEAT
		\STATE for {\bf every} $x^{(i)} \in \mathcal{F}$
		{\bf set} $c^{(i)} := arg \min\limits_{j} \mid\mid x^{(i)}-\mu_j\mid\mid$
		\STATE for {\bf each} $j$ {\bf set} $\mu_j := \frac{\sum_{i=1}^{m}1\{c^{(i)} = j\}x^{(i)}}{\sum_{i=1}^{m}1\{c^{(i)} = j\}}$
		\UNTIL {convergence}
		\STATE for each image $x_i \in \mathcal{X}$, assign $x_i$ to $\mu_k$ using Equation \ref{equation:kmeansobjective}
		\FOR {{\bf each} clusters $k_i$}
		\STATE Train a GAN with \textit{m} example $\{\eta^{(1)}, \ldots \eta^{(m)} \} $ drawn from the cluster $k_i$ and m samples $ \{z^{(1)}, \ldots, z^{(m)}\}$ drawn from noise prior $ \mathit{P_g(Z)} $ using Equation \ref{equation:ganloss}
		\STATE Generate sample images and assign \textit{sparse label smoothing distribution to the generated image}
		\ENDFOR
		\STATE Add the generated images to the training set and train a CNN using Equation \ref{equation:sls2}
	\end{algorithmic}
\end{algorithm}
%
The overall comparison of our approach SLSR with the closely related methods is summarized in Table \ref{tab:slsrproperties}. Existing strategies to label GAN-based images in person re-id include  "Pseudo label" \cite{Lee2013PseudoLabel}, LSRO \cite{zheng2017unlabeled} and "All in one" \cite{Salimans2016ImprovedTF}. SLSR and LSRO adopt smooth vector while "All in one" and "Pseudo label" adopt one hot vector. The difference is that, LSRO label contribution on pre-defined classes is the same, with a fixed and manually assigned value of $\frac{1}{k}$ while SLSR dynamically assigns label and considers their similarities. This ensures different label contribution on the pre-defined classes and accurately models practical environment settings.
\section{Framework Overview} \label{frameworkoverview}
\begin{figure*}
	\centering
	\includegraphics[scale=0.4]{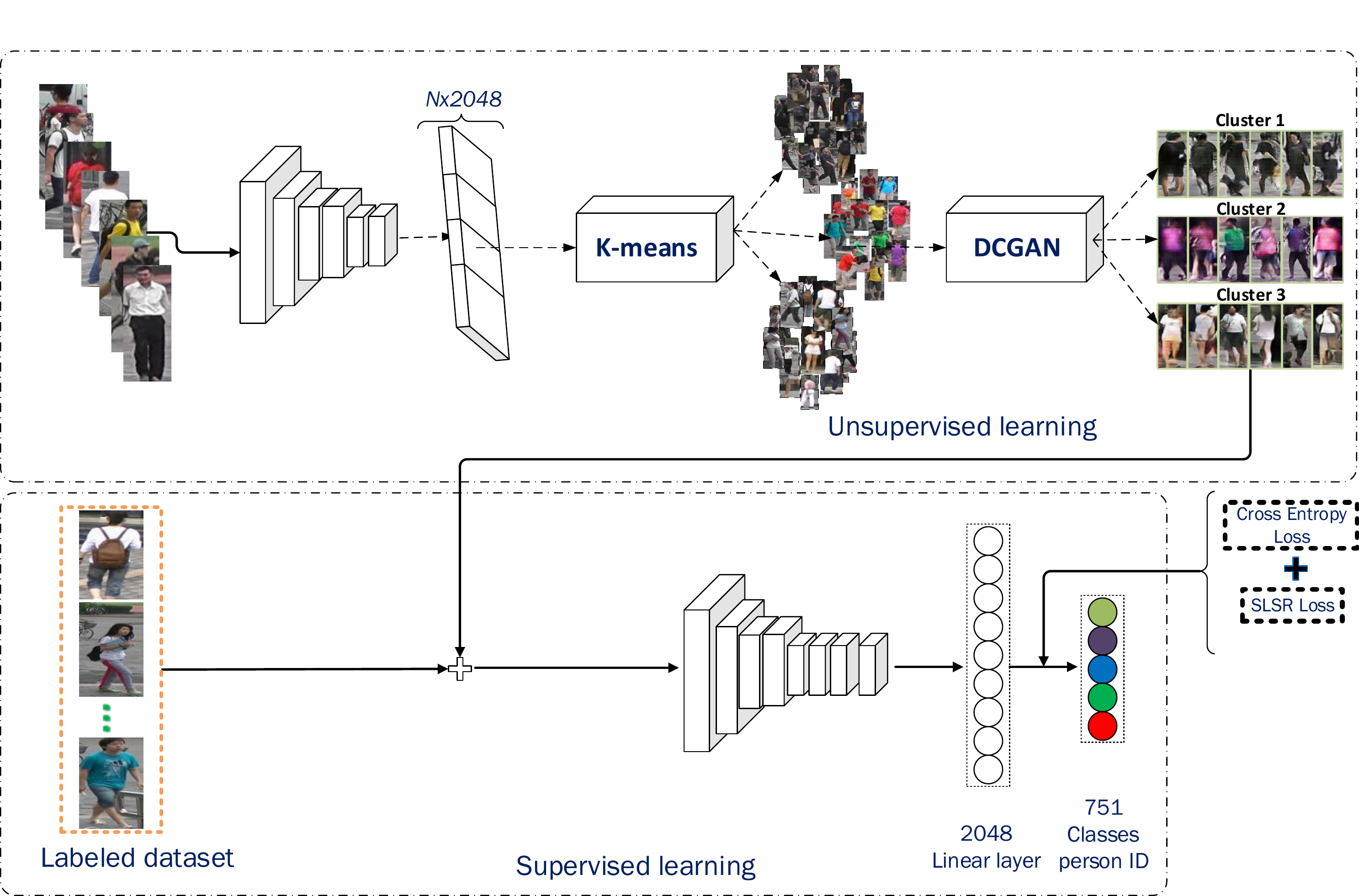}
	\caption{Our model consists of 3 steps: (1) Clustering on training data using unlabeled source dataset (Section \ref{clustering}). (2) For each cluster; train a DCGAN to generate images. Assign a partial label distribution to the generated images (Section \ref{modeling}). (3) Combine the partial labeled images with the training image.}
	\label{fig:framework}
\end{figure*}

Our framework consists of three steps as illustrated in Figure \ref{fig:framework} and includes (1) a clustering step using \textit{k-means} clustering algorithm, (2) a generative adversarial training step for image generation and finally, (3) an identity classification training task using the original training set in conjunction with the generated set.
\subsection{Clustering} \label{clustering}
\begin{figure*}
	\centering
	\includegraphics[width=\linewidth]{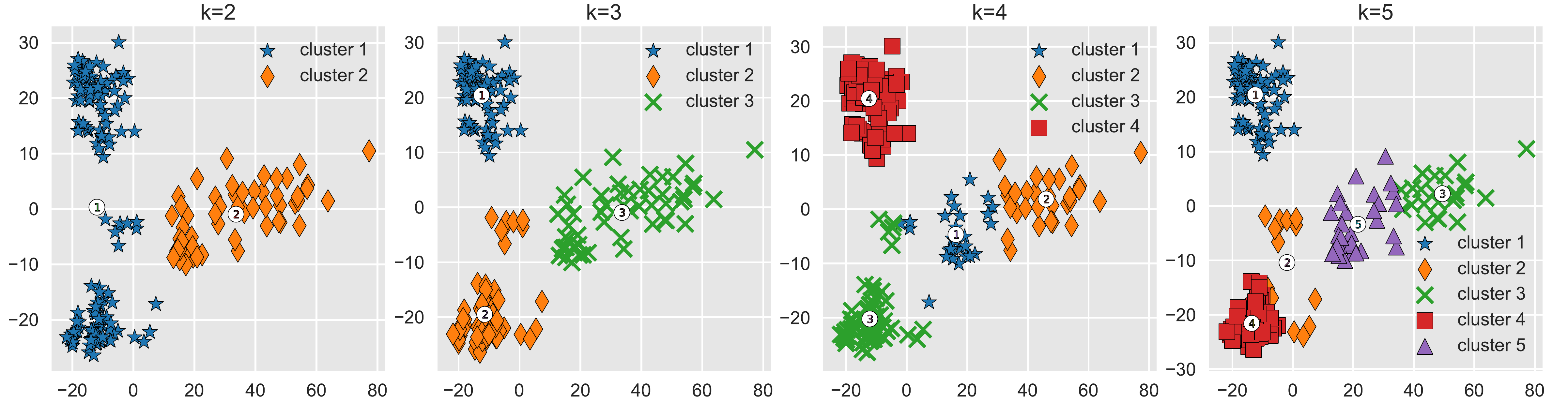}
	\caption{Visualization of extracted feature map $\mathcal{F}$ from ResNet on Market1501 dataset. Results of \textit{k-means} clustering algorithm on $\mathcal{F}$ for $k = 2, \ldots, 5$. We arrive at a fair clustering view with $k = 3$ and $k = 4$. Best viewed in color.}
	\label{fig:clusters}
\end{figure*}

It is well known that multi-view data object admits a common clustering structure across view and that person re-id is a cross-camera retrieval task across view. We aim at exploring such clustering propriety to generate images that model cross-view variations through the use of \textit{k-means} clustering algorithm and GAN. We apply \textit{k-means} algorithm to cluster the training images into \textit{K} clusters ($2, \ldots , 5$ ) as illustrated in Figure \ref{fig:clusters}. \textit{K-means} clustering is a simple yet very effective unsupervised learning algorithm for data clustering. It clusters data based on the Euclidean distance between data points.
We trained a CNN network for $40$ epochs  using a learning rate of $0.001$ with a momentum of $0.9$. We use ResNet50 \cite{Kaiming2015} model to learn a good intermediate representation and later extract high dimension features representation from the last convolutional layer. \textit{K-means} clustering algorithm is applied to the set of feature map. We found this way to be faster and better than clustering on raw data images.

To judge the effectiveness of our clustering algorithm, we considered the ground truth not known and performed an evaluation using the model itself. Table \ref{tab:silhouette} shows the cluster quality metric Silhouette Coefficient \cite{Rousseeuw1987} applied on Market-1501 dataset \cite{Zheng2015}. We found Silhouette Coefficient higher for $K=3$ and $K=4$ showing that good cluster is achieved with these values of $K$. In the next sections, we use $K=3$ for all the remaining experiments.
\begin{table}
	\caption{For each cluster size, we calculate the silhouette coefficient \cite{Rousseeuw1987} using mean intra-cluster distance (\textit{a}) and mean nearest-cluster distance (\textit{b}) $(\frac{b - a}{max(a, b)}) $. The silhouette coefficient is generally higher when clusters are dense and well separated (best value is 1 and the worse value is -1). We show that this score is higher for cluster $size = 3$. Results from Table \ref{tab:allresults} prove that we achieve higher accuracy  for $k = 3$, on Market-1501 dataset.}
	\centering
	\begin{tabular}{lc}
		\hline
		Number of clusters & Average silhouette score \\
		\hline\hline
		2 & 51.75\% \\
		\textbf{3} & \textbf{70.03\%} \\
		4 & 68.49\%\\
		5 & 61.76\%\\
		\hline
	\end{tabular}
	\label{tab:silhouette}
\end{table}
\subsection{Generative Adversarial Network}
In this second step of our framework, we used Deep Convolution Generative Adversarial Network (DCGAN) \cite{Radford2015} to generate data from clusters.  We followed the implementation details of \cite{Radford2015}. The Generator \textit{G} consists of a Deconvolutional Network (DNN) made of $8\times8\times512$ linear function, a series of four deconvolution operations with a filter size of $5\times5$ and a stride of $2$, and one $\tanh$ function. The input shape of \textit{G} is a $100$-dim uniform distribution \textit{Z} scaled in the range of $[-1,1]$ and the output shape a sample image of size $128\times128\times3$.
The Discriminator \textit{D} consists of Convolutional Neural Network (CNN) formed by four convolution functions with $5\times5$ filters and a stride of $2$.  We added a linear layer followed by a $sigmoid$ function to discriminate real images against fake images. The input shape includes sample images from \textit{G} and real images from the training set. Each convolution and deconvolution layer is followed by a batch normalization \cite{Ioffe2015} and \textit{ReLU} in both the generator and discriminator.
\subsection{Convolutional Neural Network}
In the last step of the framework, we fine-tuned the ResNet \cite{Kaiming2015} baseline model pre-trained on ImageNet, we introduced an extra linear layer into the network which adapts the network outputs to match the noisy GAN label distribution.
 The network was able to adjust the weights based on the error when we add a linear layer on top of the softmax layer rather than a non-linear such as $tanh$ or $ReLU$.
 We used the generated data in conjunction with the labeled data and defined a loss function with a regularization term. The model is trained to minimize the loss function.
\section{Experiments} \label{experiments}
In this section, we performed experiments on four widely adopted \st{and large-scale}person re-id datasets. The evaluation code is available at \href{https://github.com/jpainam/SLS\_ReID}{https://github.com/jpainam/SLS\_ReID} and is mainly conducted on Market-1501 dataset.
\subsection{Person Re-ID datasets}
Table \ref{tab:datasets} gives detailed information of the testing/training split strategy adopted during the experiments on Market-1501, CUHK03, DukeMTMC-ReID and VIPeR datasets.

{\bf Market-1501} \cite{Zheng2015} is a large and most realistic dataset collected in front of a campus supermarket. It contains overlapping views among the six cameras and images were automatically detected by the Deformable Part Model (DPM) \cite{Felzenszwalb2010}. The dataset contains $12,936$ images with $751$ identities in the training set and $19,732$ images with $750$ identities in the test set. We follow the standard data separation strategy as described in \cite{Zheng2015} and use all the training set for the clustering step and one image per identity as validation image in the last step.

{\bf CUHK03} \cite{Li2014Pairing} contains $13,164$ images and $1,467$ identities. The dataset provides two image sets, one set is automatically detected by the Deformable Part Model \cite{Felzenszwalb2010}, and the other set contains manually cropped bounding boxes. Misalignment, occlusions and body part missing are quite common in the detected set. In this work, we use the detected set as it is more realistic. The dataset is captured by six cameras, and each identity has an average of $4.8$ images in each view.

{\bf DukeMTMC-ReID} \cite{zheng2017unlabeled} is a dataset derived from the DukeMTMC \cite{Ristani2016}  dataset for multi-target tracking. The original dataset consists of a video data set recorded by $8$ synchronized cameras over $2,000$ unique identities. In this paper, we use the subset as defined by \cite{zheng2017unlabeled}. It contains $16,522$ training images with $702$ identities and $17,661$ test images with $702$ identities. We follow the partition settings of the Market-1501 dataset and use all the training images for the first step and randomly pick one image per identity as validation set. The remaining training images are used for the supervised learning step.

{\bf VIPeR}\cite{Gray2007evaluating} contains 632 pedestrian image pairs captured outdoor from two viewpoints. Each pair contains two images of the same individual cropped and scaled to $128\times 48$ pixels. The datasets are divided into two equal subsets. To be fair in the comparison, we follow the testing strategy as defined in \cite{Gray2007evaluating, Zhao2017}.
\begin{table}
	\caption{Dataset split details. The total number of images (\textit{QueryImgs, GalleryImgs, TrainImgs}), together with the total number of identities (\textit{TrainID, TestID}) are listed.}
	\centering
	\begin{tabular}{lccccc}
		\hline
		Dataset &Market&CUHK03&VIPeR&Duke\\
		\hline\hline
		\#IDs&1501&1,467&632&1404\\
		\#Images&36,036&14,097&1,264&36,411\\
		Cameras&6&2&2&8\\
		TrainID&751&1367&316&702\\
		TrainImgs&12,936&13,113&625&16,522\\
		TestID&750&100&316&702\\
		QueryImgs&3,368&984&632&2,228\\
		GalleryImgs&19,732&984&316&17,661\\
		\hline
	\end{tabular}
	\label{tab:datasets}
\end{table}

\subsection{Implementation details}
We modified ResNet50 \cite{Kaiming2015} last fully connected layer with the number of classes i.e. $751$; $1,367$ and $702$ units for Market-1501, CUHK03 and DukeMTMCReID respectively. To train the network, we used stochastic gradient descent and start with a base learning rate of $\eta^{(0)}=0.01$ and gradually decrease it as the training progresses using the inverse policy $\eta^{(i)} = \eta^{(0)} (1 + \gamma \cdot i)^{-p}$, where $\gamma = 0.1$, $p=0.025$ and $i$ is the current mini-batch iteration. We used a momentum of $\mu=0.9$ and weight decay of $\lambda=5\times10^{-4}$ and the mini-batch size of $32$. We trained the network for $130$ epochs. To generate image samples, we trained DCGAN for $30$ epoch using Adam \cite{Kingma2014Adam} with learning rate $lr=0.0002$ and $\beta_1=0.5$.

{\bf Data preprocessing}: All the input images are resized to $256\times256$ before being randomly cropped into $224\times224$ with random horizontal flip. We scaled the pixels between $-1$ and $1$. Finally, pixels are zero-centered by subtracting their mean in each dimension and random erasing \cite{Zhong2017} is applied to make the network more robust to variations and occlusions.

\begin{table*}
	\centering
	\caption{Impact of the number of cluster on Market-1501 dataset. As the number of cluster gets larger, the accuracy drops. In general, we find that a large \textit{k} decreases the training error but increases the validation/testing error. We show results of applying SLSR for $3$ different values of \text{k} with \textbf{no re-ranking}\cite{Zhong2017reranking} and single query setting. The best results are obtained with $K=3$ and $K=4$. $K=3$ is used for experiments on all the datasets}
	\label{tab:allresults}
	\resizebox{\linewidth}{!}{%
		\begin{tabular}{lcccc|cccc|cccc|ccccccc}
			\hline
			Cluster size & \multicolumn{4}{c|}{K = 2} & \multicolumn{4}{c}{K = 3} & \multicolumn{4}{|c}{K = 4}&\multicolumn{4}{|c}{K = 5}\\
			\hline\hline
			Generated &R1&R5&R10 &mAP&R1&R5&R10&mAP&R1&R5&R10&mAP&R1&R5&R10&mAP \\
			\hline
			6,000&88.30&95.90&97.50&74.08&90.59&96.58&97.86&77.56&88.92&95.93&97.62&74.31&87.32&94.00&96.31&65.88\\
			8,000&88.98&95.75&97.56&74.35&91.18&96.94&98.13&78.43&90.08&96.73&98.01&76.25&88.03&94.65&96.46&67.34\\
			12,000 &\textbf{89.99}&\textbf{96.41}&\textbf{98.04}&\textbf{75.47}&\textbf{92.43}&\textbf{97.27}&\textbf{98.39}&\textbf{79.08}&\textbf{91.36}&\textbf{97.06}&\textbf{98.22}&\textbf{79.14}&\textbf{88.48}&\textbf{95.96}&\textbf{97.56}&\textbf{73.62}\\
			18,000&89.49&96.17&97.62&75.63&91.95&96.70&98.24&78.94&91.06&96.85&98.07&78.30&88.56&95.75&97.26&74.00\\
			24,000&89.49&96.08&97.53&75.10&91.15&96.43&97.71&78.21&91.05&96.79&98.19&77.40&87.85&95.25&96.91&72.78\\
			\hline
		\end{tabular}
	}
\end{table*}
\begin{table*}
	\caption{Comparison results with LSRO using our baseline. We applied LSRO loss on our baseline on Market-1501 dataset without re-ranking. We show that the architectural design of our baseline also benefits LSRO. SQ stands for Single Query and MQ for Multi-Query.}
	\centering
	\begin{tabular}{lccccccc}
		\hline
		\multirow{2}{*}{Methods}& \multicolumn{2}{c}{Market1501 SQ}&\multicolumn{2}{c}{Market1501 MQ}&\multicolumn{2}{c}{CUHK03}\\
		\cline{2-5}
		&R1&mAP&R1&mAP&R1&mAP\\
		\hline
		Our  Baseline&87.29&69.70&91.27&76.94&75.11&83.91\\
		LSRO + Original Baseline \cite{zheng2017unlabeled}&83.97&66.07&88.42&76.10&84.62&87.40\\
		LSRO + Our Baseline &\underline{88.63} $\uparrow$4.66&\underline{74.95}$ \uparrow$8.88
&\underline{91.42}$\uparrow$ 3.00&\underline{79.87}$ \uparrow$3.77&
88.76$\uparrow$4.14&90.02$\uparrow$2.62\\
		\hline
		\textbf{SLSR}&\textbf{89.16}$\uparrow$\textbf{5.19}&\textbf{75.15}$\uparrow$\textbf{9.08}&
\textbf{92.25}$\uparrow$\textbf{3.83}&\textbf{81.92}$\uparrow$\textbf{5.82}&{\bf 91.03}&{\bf 94.21}\\
		\hline
	\end{tabular}
	\label{tab:lsrovsslsr}
\end{table*}
\subsection{Baseline models comparison}
We also compared SLSR and LSRO using our baseline. At first glance, our baseline already outperforms LSRO as it is reported in Table \ref{tab:lsrovsslsr}. Our baseline model fine-tuned ResNet model with an extra linear layer for the noisy data distribution and introduced a $512$-bottleneck layer before the softmax layer  while the baseline model used by LSRO makes no change to the existing ResNet architecture. For a fair comparison, we evaluated LSRO model on our baseline and showed the results of the experiments in Table \ref{tab:lsrovsslsr}.
For instance, on Market1501 dataset, our baseline model improves LSRO by a factor of $4.66\%$ on rank-1 accuracy and by a factor of $8.88\%$ on mAP accuracy. This shows that the architectural design of our baseline also benefits LSRO. Such baseline can be adopted to improve the overall person re-id accuracy.
Using the same baseline, we still observed a slight performance improvement. On Market-1501 dataset for example, under single query setting, SLSR slightly outperforms LSRO by a factor of $0.2\%$ on mAP accuracy and $0.53\%$ on rank-1 accuracy while under multi-query setting, SLSR outperforms LSRO by a factor of $2.05\%$ on mAP accuracy  and $0.83\%$ on rank-1 accuracy.
This improvement is explained by the relatively small size of the label distribution in Market1501 dataset. We recall that Market1501 dataset \cite{Zheng2015} contains $751$ identities for $12,936$ training images. In this case, LSRO will assign a reasonable smooth value of 0.001 (1/751) while our method with 3 clusters will assign a relative value of $0.004$. The two values are relatively closed. So, during training, the two models can converge identically.
Nonetheless, 
 in order to verify the effectiveness of the proposed method on a large class dataset and verify its robustness against the over-smoothness problem, we conducted an empirical study on CUHK03 dataset \cite{Li2014Pairing}]. As a quick reminder, CUHK03 dataset \cite{Li2014Pairing} contains $1367$ identities for $13,113$ images, making it one of the largest dataset in person re-id in term of label distribution. The comparison of the results in Table \ref{tab:lsrovsslsr} clearly shows that our model stands out from LSRO when the class label distribution is large. In details, we achieved a rank-1 accuracy improvement of $2.27\%$ and a mAP accuracy improvement of $4.19\%$. We conclude that our model can better handle practical environment scenario with thousands of labels.

\subsection{The impact of using different number of cluster}
The impact of using different numbers of clusters and different number of synthesized images during training is also evaluated and reported in Table \ref{tab:allresults}. We performed an ablation study and a performance comparison using $6000, 8000, 12000, 18000$ and $24000$ unlabeled images and expected the model to increasingly learn discriminative pattern from these data. However, the results show that as the number of generated samples increases, the person re-id performance improves by a factor of $1.25\%$ but reaches saturation with $12,000$ generated samples.
We note that the number of training images in Market-1501 dataset is $12,936$. As a result, we make two remarks. First, the addition of different numbers of fake samples steadily improves the baseline. We find that the peak performance is achieved by roughly doubling the number of training samples with fake samples. Compared with LSRO where the peak performance is achieved when $2 \times GAN$ i.e. $24,000$ images are added, our approach only requires $12,000$ to reach peak performance. Also, increasing the number of GAN images beyond $12,000$ does not improve the accuracy. The network reaches early convergence thanks to SLSR. In addition, the number of cluster affects the rank-1 accuracy. In fact, if $K=1$, the approach resembles LSRO; with $K > 2$ and $K<5$, we observe accuracy improvement over the baseline but a drop in accuracy with $K>5$. As the number of cluster increases, the learning procedure tends to converge towards assigning a single ground truth label to the fake samples similar to 'Pseudo label' scheme, which is not desirable. Therefore, we conclude that a trade-off is recommended to avoid poor regularization of partial label distribution.

\subsection{Evaluations}
 We adopted the widely used Cumulative Matching Curve (CMC) metric for quantitative evaluations.
We used the standard protocol to ensure fair comparison between the proposed method and the state-of-the-art methods. The test protocols are as follow.

For VIPeR dataset, we randomly divide the dataset into training and testing sets, each set containing half of the available individuals. In the test set, we randomly select one image of a person from camera 1 as a query image and one image of the same person from camera 2 as a gallery image.  For CUHK03 dataset, we followed the standard protocol used by \cite{Chen2018Correlation} and for Market-1501 dataset, we used the standard evaluation protocol as defined by \cite{Zheng2015}. And, for DukeMTMC-ReID we used the standard evaluation protocol defined in \cite{zheng2017unlabeled}.
Both single-query and multi-query matching results are reported on Market-1501 dataset while only single query evaluation is adopted for CUHK03, VIPeR and  DukeMTMC-ReID datasets. Rank-1, rank-5, rank-20 accuracy and Mean Average Precision (mAP) are computed to evaluate the performance of all the methods.
For each image in the query set, we first compute the  L2 distance between the query image and all the gallery images using the output feature produced by our trained network, and we return the top-n nearest images in the gallery set. If the returned list contains an image of the same person at a given position $k$, then this query is considered as success at rank-k.

{\bf Re-ranking}: Recent works \cite{Bai2017, Zhong2017reranking} choose to perform an additional re-ranking to improve ReID accuracy. In this work, we report re-ranking results using re-ranking with \textit{k}-reciprocal encoding \cite{Zhong2017reranking}, which combines the original L2 distance and Jaccard distance. Re-ranking with  \textit{k}-reciprocal encoding approach assumes that there are multi positive samples in the gallery. So, re-ranking approach will fail to improve the performance in small datasets such as ViPER and CUHK03 datasets. In this work, we did not report these results. In Tables \ref{tab:cuhk03results} \ref{tab:dukeresults} \ref{tab:marketresults} \ref{tab:viperresults}, SLSR represents our method and SLSR+RR represents our model with re-ranking \cite{Zhong2017reranking}.

\begin{table}
	\caption{Comparison result with state-of-arts on CUHK03. '-' means that no reported results is available. * paper on ArXiv but not published}
	\label{tab:cuhk03results}
	\centering
	\resizebox{\linewidth}{!}{%
	\begin{tabular}{lcccc}
		\hline
		Methods&R1&R5&R10&mAP \\
		\hline \hline
		KISSME \cite{Kostinger2012} & 11.7&33.3&48.0&-\\
		DeepReID \cite{Li2014Pairing}&19.89&50.00&64.00&-\\
        TAUDL \cite{Minxian2018Tracklet}&44.7 &&&31.2\\
		ImprovedDeep \cite{Ahmed2015}&44.96&76.01&83.47&-\\
		XQDA (LOMO) \cite{Liao2015} &46.25&78.90&88.55&-\\
		SI-CI \cite{Wang2016}&52.20&84.30&94.8&-\\
		DNS \cite{Zhang2016}	&54.7&80.1&88.30&-\\
		FisherNet \cite{Wu2016FisherNet}&63.23&89.95&92.73&44.11\\
		MR B-CNN \cite{Ustinova2015}&63.67&89.15&94.66&-\\
		Gated ReID \cite{Varior2016Gated}&68.1&88.1& 94.6&58.8\\
		SOMAnet \cite{Barros2018}&72.40&92.10&95.80&-\\
		SSM \cite{Bai2017}&72.7&92.4&96.1&-\\
		SVDNet \cite{Sun2017SVDNet}&81.8&95.2&97.2&84.8\\
		Cross-GAN \cite{Zhang2018Crossing}*&83.23&-&96.73&-\\
		Verif.Identif. \cite{zheng2016discriminatively}&83.40&97.10&98.7&86.40\\
		DeepTransfer \cite{Geng2016}*&84.10&-&-&-\\
		LSRO \cite{zheng2017unlabeled}&84.62&97.60&98.90&87.40\\
		TriNet \cite{Hermans2017Triplet}&87.58&98.17&-&-\\
		HydraPlus-Net \cite{liu2017hydraplus}&\textcolor{blue}{\bf91.8}&\textcolor{blue}{\bf98.4}&99.1&-\\
		\hline \hline			
		{\bf (Ours)} SLSR&{\bf91.03}&{\bf98.22}&{\bf99.26}&{\bf 94.21}\\	
		\hline
	\end{tabular}
	}
	
\end{table}
\begin{table}
	\caption{Comparison results of the state-of-arts methods on DukeMTMCReID. We show that our methods is superior to previous works. * paper on ArXiv but not published}
	\label{tab:dukeresults}
	\centering
	\resizebox{\linewidth}{!}{%
		\begin{tabular}{lcccc}
			\hline
			Methods&R1&R5&R10&mAP \\
			\hline \hline
			BoW+KISSME \cite{Zheng2015}&25.13&-&-&12.17\\
			XQDA (LOMO) \cite{Liao2015}&30.75&-&-&17.04\\
            TAUDL \cite{Minxian2018Tracklet}&61.7&&& 43.5\\
			LSRO \cite{zheng2017unlabeled}&67.68&-&-&47.13\\
			OIM \cite{xiaoli2017joint}&68.1&-&-&47.4\\
			TriNet \cite{Hermans2017Triplet}*&72.44&-&-&53.50\\
			SVDNet\cite{Sun2017SVDNet}&\textcolor{blue}{\bf76.7}&86.4&89.9&56.8\\
			\hline \hline
			{\bf (Ours)} SLSR&{\bf76.53}&{\bf88.15}&{\bf 91.02}&{\bf60.79}\\
			{\bf (Ours)} SLSR+RR&\textbf{82.67}&\textbf{89.72}&\textbf{93.00}&\textbf{79.23}\\
			\hline
		\end{tabular}
	}
	
\end{table}
\begin{table}[t]
	\caption{Comparison results of the state-of-art methods on Market-1501. '-' means that no reported results is available  and '*' means the paper is available on ArXiv but not published}
	\label{tab:marketresults}
	\centering
	\resizebox{\linewidth}{!}{%
		\begin{tabular}{lcccc}
			\hline
			& \multicolumn{4}{c}{Single Query}\\
			\hline
			Methods&R1&R5&R10&mAP \\
			\hline
			BoW+KISSME \cite{Zheng2015}&44.42&-&-&20.76\\
			FisherNet \cite{Wu2016FisherNet}&48.15&-&-&29.94\\
			Simil.Learning \cite{Chen2016}&51.90&-&-&26.35\\
			DNS \cite{Zhang2016}&61.02&-&-&35.68\\
            TAUDL \cite{Minxian2018Tracklet}&63.7&&& 41.2\\
			Gate Reid \cite{Varior2016Gated}&65.88&-&-&39.55\\
			MR B-CNN \cite{Ustinova2015}&66.36&85.01&90.17&41.17\\
			Cross-GAN \cite{Zhang2018Crossing}*&72.15&-&94.3&48.24\\
			SOMAnet \cite{Barros2018}&73.87&88.03&92.22&47.89\\
			HydraPlus-Net \cite{liu2017hydraplus}&76.9&91.3&94.5&-\\
			Verif.Identif \cite{zheng2016discriminatively}&79.51&-&-&59.87\\
			SVDNet \cite{Sun2017SVDNet}&82.3&92.3&95.2&62.1\\
			DeepTransfer \cite{Geng2016}*&83.7&-&-&65.5\\
			LSRO \cite{zheng2017unlabeled}&83.97&-&-&66.07\\
			TGP-ReID \cite{Almazan2018}*&92.2&97.9&-&81.2\\
			\hline \hline
			{\bf(Ours)} SLSR&{\bf89.16}&{\bf95.78}&{\bf97.33}&{\bf75.15}\\	
			{\bf(Ours)} SLSR+RR&\textbf{91.54}&\textbf{95.37}&\textbf{96.62}&\textbf{88.09}\\
			\hline \hline
			& \multicolumn{4}{c}{Multi Query}\\
			\hline
			Methods&R1&R5&R10&mAP \\
			\hline
			DNS \cite{Zhang2016}&71.56&-&-&46.03\\
			Gate Reid \cite{Varior2016Gated}&76.04&-&-&48.45\\
			SOMAnet \cite{Barros2018}&81.29&92.61&95.31&56.98\\
			Verif.Identif \cite{zheng2016discriminatively}&85.47&-&-&70.33\\
			LSRO \cite{zheng2017unlabeled}&88.42&-&-&76.10\\
			DeepTransfer \cite{Geng2016}*&89.6&-&-&73.80\\
			TGP-ReID \cite{Almazan2018}*&94.7&98.6&-&87.3\\
			\hline \hline	
			{\bf(Ours)} SLSR&{\bf92.25}&{\bf97.51}&{\bf 98.34}&{\bf 81.92}\\
            {\bf(Ours)} SLSR+RR&{\bf94.18}&{\bf98.06}&{\bf 98.78}&{\bf 90.10}\\					
			\hline
		\end{tabular}
	}
	
\end{table}

\begin{table}
	\caption{Comparison results with state-of-arts on VIPeR dataset.}
	\label{tab:viperresults}
	\centering
	\resizebox{\linewidth}{!}{%
	\begin{tabular}{lcccc}
		\hline
		Methods&R1&R5&R10&R20 \\
		\hline \hline
		ImproveDeep \cite{Ahmed2015}&34.81&63.61&75.63&84.49\\
		KISSME \cite{Kostinger2012}&34.81&60.44&77.22&86.71\\
		Simil.Learning \cite{Chen2016}&36.80&70.40&83.70&91.70\\
		MFA (LOMO)\cite{Xiong2014}&38.67&69.18&80.47&89.02\\
		XQDA (LOMO) \cite{Liao2015}&40.00&68.13&80.51&91.08\\
		Cross-GAN \cite{Zhang2018Crossing}*&49.28&-&\textcolor{blue}{\bf 91.66}&93.47\\
		DNS \cite{Zhang2016}&51.17&\textcolor{blue}{\bf 82.09}&90.51&95.92\\
		SSM \cite{Bai2017}&53.73&-&91.49&\textcolor{blue}{\bf 96.08}\\
		SpindleNet \cite{Zhao2017}&53.80&74.1&83.2&92.1\\
		HydraPlus-Net \cite{liu2017hydraplus}&56.6&78.8&87.0&92.4\\
		\hline \hline	
		{\bf (Ours)} SLSR	&{\bf 65.98}&\textbf{81.49}&\textbf{88.45}&\textbf{95.25}\\
		\hline
	\end{tabular}
	}
	
\end{table}


\subsection{Comparison with the state of art}
In this section, we compare our results with state-of-art methods and report the results in Tables  \ref{tab:cuhk03results} \ref{tab:dukeresults} \ref{tab:marketresults} \ref{tab:viperresults}.

On \textbf{Market-1501} dataset our method achieved an {\bf 89.16\%} rank-1 accuracy and {\bf 75.15\%} mAP accuracy exceeding LSRO \cite{zheng2017unlabeled} by a factor of {\bf 5.19\%} on rank-1 accuracy and by a factor of {\bf 9.08\%} on mAP accuracy. Our method with both SLSR and re-ranking \cite{Zhong2017reranking} with \textit{k}-reciprocal encoding further improves rank-1 and mAP accuracy from 89.16\% to {\bf 91.54\%} and from 75.15\% to {\bf 88.09\%} respectively. Table \ref{tab:marketresults} shows that our method outperforms many existing works. 

On \textbf{CUHK03} dataset (Table \ref{tab:cuhk03results}), we achieved a {\bf 91.03\%} rank-1 accuracy and {\bf 94.21\%} mAP accuracy which are close by a factor of {\bf 0.77\%} to the result reported by HydraPlus-Net \cite{liu2017hydraplus}. Our method exceeds LSRO \cite{zheng2017unlabeled} by a factor of {\bf 6.41\%} on rank-1 accuracy and by a factor of {\bf 6.81\%} on mAP.

Not many reported results exist  on \textbf{DukeMTMCReID} dataset,  as shown in Table \ref{tab:dukeresults}. Yet, our method achieved a {\bf 76.53\%} rank-1 accuracy and {\bf 60.79\%} mAP accuracy exceeding existing works. Compared to LSRO \cite{zheng2017unlabeled}, our rank-1 accuracy exceeds their result by a factor of {\bf 8.85\%}. SVDNet \cite{Sun2017SVDNet} exceeds our model by a small factor of {\bf 0.17\%}.

We also achieved competitive result on a small dataset such as \textbf{VIPeR} dataset, Specifically, our method achieved a {\bf 65.98\%} rank 1 accuracy.  

\begin{figure}
	\centering
	\includegraphics[width=0.8\linewidth]{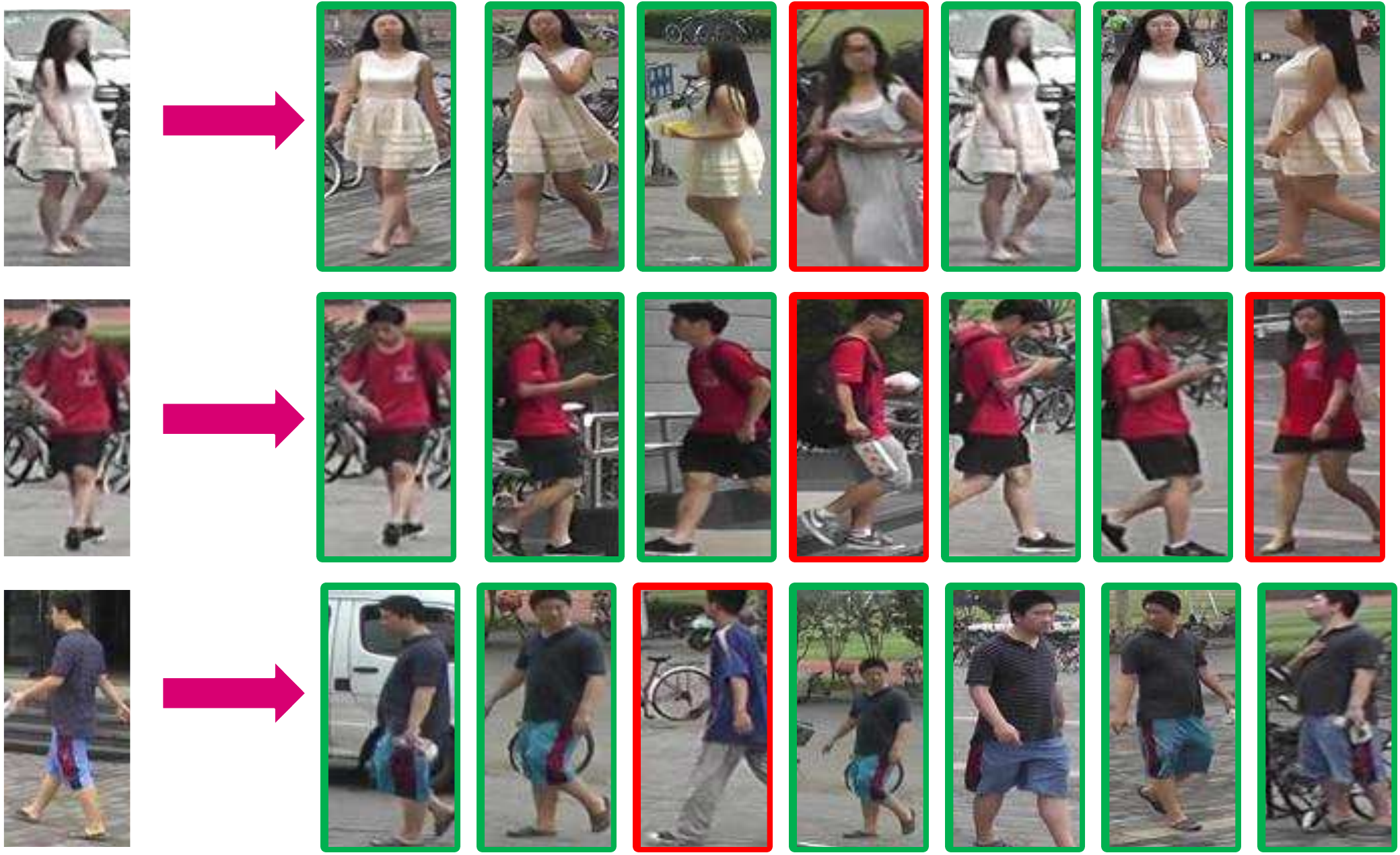}
	\caption{Sample images retrieved from Market-1501 dataset using our framework. The images in the first column are the query images. The images in the right columns are the retrieved images. The retrieved images are sorted according to the similarity scores from left to right. We use re-ranking \cite{Zhong2017reranking} with \textit{k}-reciprocal encoding.}
	\label{fig:imageretrieval}
\end{figure}


\section{Conclusion} \label{conclusion}
In this paper, we proposed Sparse Label Smoothing Regularization  (SLSR) for solving the person re-identification problem. We proposed to use generated samples in conjunction with training samples to improve the re-id accuracy and proposed a labeling approach for generated samples. We emphasized on the fact that a fair labeling approach on synthesized images should consider the underlying relationship between the training and the generated samples. We proposed SLSR as a pipeline to train a CNN model with labeled and synthesized images. We clustered the  training images using an intermediary feature representation of a pre-trained CNN model and generate images for each cluster. The generated images are assigned smooth label according to the label distribution of the cluster used for DCGAN stream. Through ablation, we show that SLRS can address the problem of over-smoothness found in current regularization methods. 
Extensive evaluations were conducted on four large-scale datasets to validate the advantage of the proposed model on existing models. Tables \ref{tab:cuhk03results} \ref{tab:dukeresults}   \ref{tab:marketresults} \ref{tab:viperresults} show the superiority of the model over a wide variety of state-of-art methods.

\section{Acknowledgements}
This work is supported by the Ministry of Science and Technology of Sichuan province (Grant No. 2017JY0073) and Fundamental Research Funds for the Central Universities in China (Grant No. ZYGX2016J083). We appreciate Yongsheng Peng, Eldad Antwi-Bekoe for their useful contributions and Yuyang Zhou for the management of the GPUs during experiments.
	{\small
		\bibliographystyle{ieee}
		\bibliography{egbib}
	}
	
\end{document}